\documentclass[runningheads]{llncs}
\usepackage[T1]{fontenc}
\usepackage{algorithm}
\usepackage{algpseudocode}
\usepackage{booktabs}
\usepackage{subfig}
\usepackage{graphicx}

\begin{document}

\title{Improving Online Bagging for Complex Imbalanced Data Streams}

\titlerunning{Improving Online Bagging for Complex Imbalanced Data Streams}
\author{Bartosz Przyby\l{} \and Jerzy Stefanowski}

\authorrunning{B. Przyby\l{}, J.Stefanowski}

\institute{Institute of Computing Science, Pozna\'n University of
Technology, 60-965~Pozna\'n,~Poland }

\maketitle

\begin{abstract}

Learning classifiers from imbalanced and concept drifting data streams is still a challenge.  Most of the current proposals focus on taking into account changes in the global imbalance ratio only and ignore the local difficulty factors, such as the minority class decomposition into sub-concepts and the presence of unsafe types of examples (borderline or rare ones). As the above factors present in the stream may deteriorate the performance of popular online classifiers, we propose extensions of resampling online bagging, namely Neighbourhood Undersampling or Oversampling Online Bagging to take better account of the presence of unsafe minority examples. The performed computational experiments with synthetic complex imbalanced data streams have shown their advantage over earlier variants of online bagging resampling ensembles.

\end{abstract}
\section{Introduction}
\label{introduction}

Although learning classifiers from concept drifting data streams has been intensively studied in the last decades it is still the subject of many new studies, including better consideration of various types of concept drifts or other data complexity \cite{PolikarSurvey,brzezinski2021impact}. This applies in particular to class imbalances in the stream, which also occur in some practical applications \cite{aguiar2023survey,chen2024survey}.

However, the current research on imbalanced and concept drifting streams is still not as developed as in the case of separately considered static data or streams. Moreover, the existing works mostly focus on re-balancing classes and reacting to changes caused by the \textit{varying global imbalance ratio}. These works do not sufficiently consider more complex imbalanced stream scenarios, where these changes are additionally accompanied by the \textit{local difficulty factors} already considered for the static imbalanced data such as the minority class split into sub-concepts, class overlapping, or occurrence of different types of unsafe minority examples (borderline, rare or outliers \cite{Krysia_2016}). In evolving data streams these factors could also influence the changes in \textit{local class distributions} and lead to distinguishing new types of  \textit{local drifts} in the stream \cite{brzezinski2021impact,BrzezStef2019}.

 
 Up to now, only two experimental studies have been conducted on the role of these local difficulty factors combined with other types of concept drifts \cite{brzezinski2021impact,Lipska2022}. The authors of  \cite{brzezinski2021impact} carried out a large series of comprehensive experiments with synthetic and real data streams showing the different influence of types of minority examples and class decomposition on predictions of representative online classifiers. They showed that especially re-sampling generalizations of online bagging coped sufficiently well with changes of the global class imbalance ratio in the stream and partially with the split of the minority class into smaller sub-concepts. However, drift associated with the greater presence of unsafe examples (especially rare examples) caused a deterioration in the performance of all tested online classifiers. Furthermore, combinations of multiple factors were demonstrated to be the most challenging for all classifiers, which were not able to recover from these drifts \cite{brzezinski2021impact}. The experiments of \cite{brzezinski2021impact}  concerned binary imbalanced classes, however later their main observations were confirmed also for multiple classes in \cite{Lipska2022}.

Therefore, the aim of this work is to propose new generalizations of under-sampling or over-sampling online bagging ensembles that would better handle the drifts of unsafe types of examples in such complex imbalanced data streams.

\section{Related Works} 
\label{sec:related}

Due to the page limits the reader is referred to works such as \cite{krawczyk2017} for a review of research on data streams or to \cite{Branco,chen2024survey,imb-chapter-book} for imbalanced data. In the case of imbalanced and concept-drifting data streams, refer to  \cite{aguiar2023survey,BrzezStef2019,brzezinski2021impact}.

 In the case of static imbalanced data, it has been already shown that besides the global imbalance between classes other sources of classifier deterioration include \cite{eswa2022}: (1) the decomposition of the minority class into several sub-concepts, (2) the presence of small, isolated groups of minority examples located deeply inside the majority class region (it corresponds to \textit{rare cases}), (3) the effect of strong overlapping between the classes.  

\begin{figure}[ht]
    \centering
\includegraphics[width=11cm]{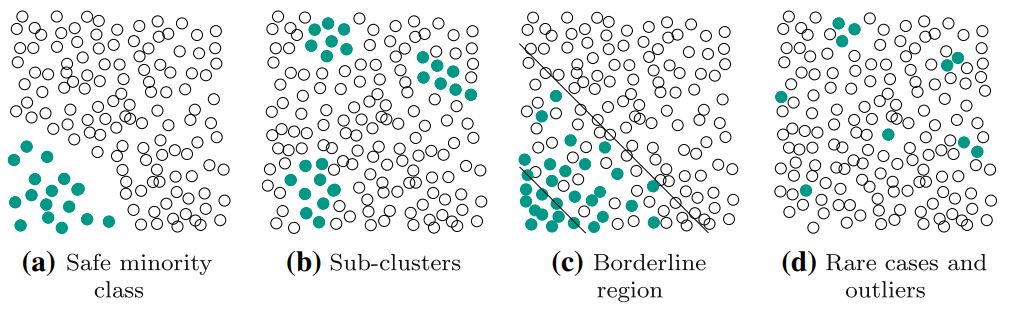}    
    \caption{An illustration of different difficulty factors in imbalanced data - based on \cite{BrzezStef2019}}
 \label{Fig:imbalance}    
\end{figure}

The last two factors can be identified through, the so-called, \textit{types of examples} \cite{Krysia_2016}, which distinguish between safe and unsafe examples. \textit{Safe examples} are the ones located in homogeneous regions populated by examples from one class only. The \textit{unsafe}  examples are categorized into \textit{borderline}~(placed close to the decision boundary between classes), \textit{rare cases}~(isolated groups of few examples located deeper inside the opposite class), and \textit{outliers} - singletons. See Figure \ref{Fig:imbalance} for their illustration.  Following the method from \cite{Krysia_2016} these types of examples can be identified based on the analysis of class labels of other examples in the local neighbourhood of the given instance.  Most studies on imbalanced data streams focus on handling either static or drifting imbalance ratios only and they do not consider the aforementioned data factors.

The authors \cite{brzezinski2021impact} introduced an \textit{extended concept drift categorization from imbalanced streams} that takes into account these local data factors, and covers specific types of drifts inside them.  These authors prepared a special generator of the synthetic data streams, where various scenarios of the occurrence of these elements are modeled - see its description in an electronic appendix of \cite{brzezinski2021impact}.

 It was observed in \cite{brzezinski2021impact} that specialized imbalanced stream classifiers (ie. two re-sampling versions of online bagging, i.e. over-sampling the minority class OOB or under-sampling the majority class UOB \cite{WangMY15}, and a specialized neural network ESOS-ELM) coped well with nearly all static and dynamically changing imbalance ratios. Non-specialized online classifiers (such as OB and VFDT) performed much worse, especially when the minority class ratio dropped down to 1--5\%.  The rest of the experiments demonstrated that the imbalance ratio can play an important (amplifying) role only when combined with other factors. Then, the other aspects of drift locality and minority class composition affected all the analyzed classifiers. All classifiers suffered moderate performance drops when exposed to moving sub-clusters and even more substantial performance drops and limited recovery when exposed to minority class splits. However, the next experiments demonstrated that changes in the distribution of types of minority examples from safe to unsafe ones (i.e. borderline or rare) were very influential, in particular for introducing more rare examples as a result of the drift. Drifting proportions of rare examples were the only drift the best classifiers did not recover from in any way. Furthermore, the effects of all the analyzed difficulty factors are amplified when they occur together as combined \textit{multiple drifts}. 

These observations for binary imbalanced classes were confirmed in a similar study for multi-class data \cite{Lipska2022}, which also showed a greater impact of the presence of borderline examples. 


\section{Neighbourhood Online Bagging}
\label{sec:method}

In our work, we decided to extend under or over-sampling ensemble approaches based on online bagging (originally proposed in \cite{WangMY15}) because they performed best in experiments with difficult imbalanced data streams \cite{brzezinski2021impact,Lipska2022}. 

Online bagging \cite{Oza}, in contrast to the static Breiman’s bagging ensemble, is capable of dealing with online environments because it processes each example only once upon arrival. The idea of online bagging is based on exploiting the Poisson distribution and its parameter $\lambda$ to estimate how many times the current incoming example should be sent to each component classifier (a default $\lambda$ = 1, which works well for approximately balanced classes in the stream). Each classifier can incrementally update (trained) with such received examples depending on this estimate.

\textit{Oversampling-based Online Bagging} (OOB) and \textit{Undersampling-based Online Bagging}  (UOB) are generalizations proposed in \cite{MUOB,WangMY15} to change the presence of examples from the particular class in this Poisson distribution operation concerning the current imbalance ratio in the stream. 

The number of examples from each class in the given time moment in the stream is continuously updated and used to calculate $\lambda$ as a function of the current imbalanced ratios. The value in the nominator of $\lambda$ depends on whether a classifier should over- or under-sampled for a currently considered concept present in the stream. In OOB the value of $\lambda$ is equal to the cardinality of the current biggest class size divided by the class size of the currently processed example. It increases the number of minority class examples (as $\lambda > $ 1) and does not change the number of majority ones $\lambda$ = 1. In the UOB the value of $\lambda$ is computed as the ratio of the current size of the smallest class in the stream and the class size of the current example. Practically, it ($\lambda$ < 1) reduces the number of majority examples sent to component classifiers. 

To sum up, the both re-sampling generalizations modify sending an incoming example from a specific class for updating component classifiers in online bagging with respect to the imbalance ratio between classes in the stream.

In our current proposal, we modify the $\lambda$ coefficient by incorporating information about the difficulty of the incoming example. Previously \cite{blaszczynski2015neighbourhood} in order to estimate the difficulty of the example in the static data,  the local analysis of class labels of its $k$ nearest neighbours was exploited. Following this inspiration we propose to define \textit{unsafeness level} of a minority class example $x$ as $$L^2_{min} = \frac{(N_{maj}')^\Psi}{k}$$
where $N_{maj}'$ is the number of examples belonging to the majority class among $k$ nearest neighbours of $x$ which are calculated on a sliding window in the stream, $k$ is the number of nearest neighbours taken for the analysis, $\Psi$ is a parameter responsible for additional amplification of the impact of unsafe examples (i.e. value $\Psi$ > 1 amplifies the role of rare cases and outliers). In the \textit{Neighbourhood Oversampling Online Bagging} (NOOB) this coefficient is aggregated with the class sizes coefficient to increase the number of unsafe minority examples send to update the component classifiers, which is defined as follows:
$$\lambda = (N_{maj}/N_{min}) \cdot (L^2_{min} + 1)$$
where $N_{maj}$ denotes the number of examples from the majority class contained in the sliding window, $N_{min}$ denotes the number of examples from the minority class. 

In (NOOB) it increases the Poisson distribution estimate of the number of minority examples while for the incoming majority examples $\lambda$ = 1 (as in the standard online bagging). The general pseudocode of the Neighbourhood Oversampling Online Bagging is presented in Algorithm  \ref{Algorithm:NOOB}).

\begin{algorithm}
    \caption{Neighbourhood Oversampling Online Bagging (NOOB)}\label{Algorithm:NOOB}
    \textbf{Input}: $S$: stream of examples \\
    \hspace*{10mm} $n$: number of classifiers in ensemble \\
    \hspace*{10mm} $W$: window of examples \\
    \hspace*{10mm} $k$: number of nearest neighbours \\
    \hspace*{10mm} $\Psi$: additional coefficient for calculating safe level \\
    \textbf{Output}: $\mathcal{E}$: an ensemble of classifiers \\
    \begin{algorithmic}[1]
    \ForAll{examples $x \in S$ \do}
    \State calculate safe level of incoming example $L^2_{min} = \frac{(N_{maj}')^\Psi}{k}$
    \If{$x \in minority\ class$}
    \vspace{0.5em}
    \State $\lambda \gets (N_{maj}/N_{min}) \cdot (L^2_{min} + 1)$
    \vspace{0.5em}
    \Else
    \State $\lambda \gets 1$
    \EndIf
    \ForAll{classifiers $C_i \in \mathcal{E}$ \do}
    \State set $l$ $\sim Poisson(\lambda)$
    \State update $C_i$ using x, l times
    \EndFor
    \State $W \gets W \cup \{x\}$
    \State if necessary remove outdated examples from $W$
    \EndFor
    \end{algorithmic}
\end{algorithm}

In similar way, we propose \textit{the Neighbourhood Undersampling Online Bagging} (NUOB), where for the incoming majority examples we define a \textit{safeness level} as $$L^2_{maj} = \frac{N_{maj}}{k}$$
 And as its consequence a new coefficient for the majority example $$\lambda = (N_{min}/N_{maj}) \cdot (L^2_{maj})^{\Psi}$$ which reduces the chance of using unsafe majority examples to update component classifiers and having $\lambda = 1 $ for incoming minority examples it leads to such undersampling which removes rare cases, outliers and partly borderline majority examples \footnote{Its pseudocode is provided in the  online appendix to this paper, "An appendix to Improving Online Bagging for Complex Imbalanced Data Streams", which is available at \url{http://www.cs.put.poznan.pl/jstefanowski/pub/imbstreams.pdf}}.
 
 Finally, as observed in experiments sometimes NUOB is better than NOOB and sometimes, on the contrary, NOOB is better than NUOB, we decided to propose their hybrid version. 
 
 In \textit{Hybrid Neighbourhood Online Bagging} (HNOB), both ensembles NUOB and NOOB are trained in parallel for each incoming example and their classification performance is continuously assessed using an evaluation metric suitable for class imbalanced data (in our experiment \textit{G--mean} of both classes). The ensemble that had a better evaluation measure value in the last evaluation is selected to make class prediction for the current example in the stream.

\section{Experiments}
\label{sec:experiments}

The aim of the experiment is to evaluate the usefulness of the proposed Neighbourhood Online Bagging ensembles and their reaction to considered data difficulty factors and drifts, in particular concerning borderline and rare minority examples. This is why we compare the 6 following classifiers:   \textit{basic online bagging, undersampling online bagging, oversampling online bagging, neighbourhood undersampling online bagging, neighbourhood oversampling online bagging, and hybrid neighbourhood online bagging}. 

We apply the same set of hyperparameters as in \cite{brzezinski2021impact} (as it was already tuning for considered synthetic data streams). So, all bagging ensembles contain 15 component Hoeffdings' trees (with their standard parameters). UOB and OOB use the forgetting degree ($\theta$ = 0.9). For all neighbourhood variants, we compute the (un)safeness levels with $k$ = 5 neighbours over the last sliding window $W$ of size equal to 500 examples. These values were found by a grid search ($k$ = 5, 7,9 and $W$ = 500 till 2000) in preliminary a experiment. Similarly, we tune the exponent degree $\Psi$ to 2.0, which means that we pay more attention to rare examples than to safe ones.

We carried out the experiments in a controlled framework based on synthetic generated data streams, where each data factor can be modeled and parametrized according to different planned scenarios.  We use a generator designed and used for the earlier related experiments from \cite{brzezinski2021impact}.  Our experiments cover the following difficulty factors and drifts:
\begin{itemize}
 \item \textit{Imbalanced ratios} -- into two scenarios of either static data or with a single drift with changing the imbalances ratio from 1\% to fully balanced classes. 
 \item \textit{Types of minority examples} being either \textit{borderline} (class overlap) or \textit{rare} ones (other examples in the minority classes are generated to be safe ones), i.e. with the given percentages of their occurrence in each class: 20\%, 40\%, 60\%, 80\%, and 100\%.
 \item \textit{Changes in class composition}, i.e. local drifts of the split of each minority class into 3, 5 and 7 sub-clusters; or merging them and moving these numbers of sub-clusters. 
 \end{itemize} 
Besides modeling the single factor in the given stream, we considered their pairs and occurrences of multiple factors, which previously   \cite{brzezinski2021impact} were identified as the most demanding.
 
Due to the high number of experiments and the limited number of pages, we only include the most important, representative results and use summaries of many comparative experiments using the non-parametric Friedman tests \footnote{Additional results are provided in the  online appendix to this paper, "An appendix to Improving Online Bagging for Complex Imbalanced Data Streams", which is available at \url{http://www.cs.put.poznan.pl/jstefanowski/pub/imbstreams.pdf}}.
 
Firstly, we examined the impact of \textbf{single factors}. As expected the static global imbalanced ratio or the drift of this ratio was well coped by all compared versions of the online bagging ensemble, except the standard online bagging for the stronger imbalance ratios 1\% and 2\%. Let's recall that both OOB and UOB worked well for such scenarios in  \cite{brzezinski2021impact,Lipska2022}. Moreover, our neighbourhood proposals are also not exploiting their additional mechanisms if the data streams do not contain other difficulty factors, so may work similarly to OOB or UOB (what was observed in our results).
 
More interesting was studying the impact of the minority class decomposition. For the minority class split into smaller sub-concepts NUOB was the best classifier with respect to Recall, while NOOB worked better than other compared classifiers (also for moving sub-concept) but with respect to G-mean. 

\begin{figure}[ht]
    \centering
    \subfloat{\includegraphics[width=5.7cm]{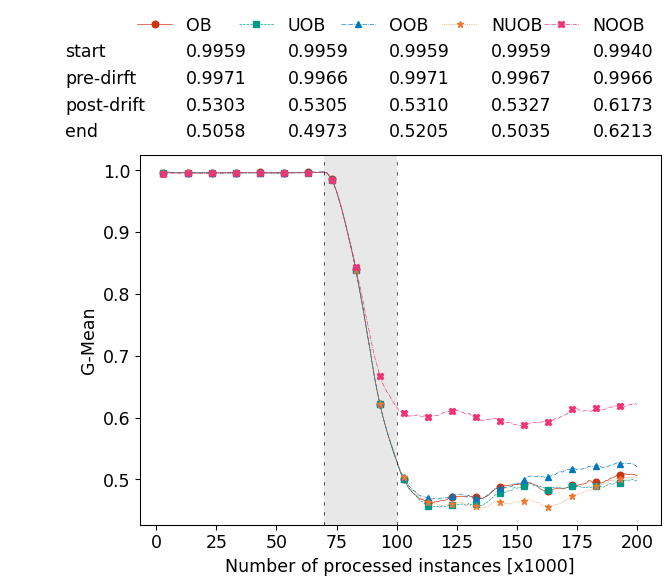}}
    \qquad
    \subfloat{\includegraphics[width=5.7cm]{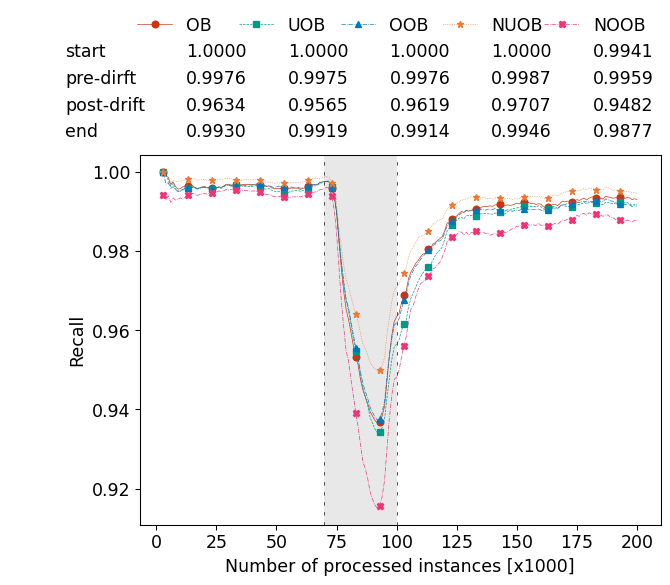}}
    \caption{Plots showing bagging variants reacting to two kinds of drift \textit{80\% rare minority examples and G-mean} measure (the left-hand figure) and  \textit{minority class split into 5 sub-concepts and Recall} measure (the right-hand figure)}
\label{Fig:single}    
\end{figure}

However, the most important experiments included data streams with different proportions of borderline and rare examples or their drifts. For both static and drift cases we noticed better performance of the new proposed neighbourhood baggings in comparison to earlier UOB and OOB ones. In general, NOOB was the best classifier for the higher proportions of rare minority examples, while NUOB worked better for borderline examples. Exemplary reactions to such drifts are shown in Figures \ref{Fig:single}.

In Table \ref{Tab:SingleDriftFriedman} we summarize results of Friedman tests carried out over aggregated results for single elements (static factors and drifts) for considered sub-categories concerning Recall measure. One can easily notice that nearly for all categories the proposed NUOB is either the best classifier or one of the best classifiers. In particular, for rare cases or borderline it is superior to the other classifiers.  On the other hand, for G-mean measure performance NOOB is better (which indicates achieving a better balance of recognizing correctly both minority and majority classes) - in particular for different proportions of rare examples NOOB followed by OOB are the best classifiers (averaged ranks in the Friedman test are 1.9 and 2.1 respectively, while NUOB and UOB have 4.0 and 4.2 -- for the critical difference \textit{CD} = 2.01 -- it is a significant dominance of NOOB.

\begin{table}[ht]
\centering\small%
\setlength{\tabcolsep}{10pt} 
\renewcommand{\arraystretch}{1.5} 
\begin{tabular}{l c c c c c c}
\toprule
Data factor &  OB & UOB & OOB & NUOB & NOOB & CD \\
\midrule
Static imbalance & 4.35 & \textbf{2.41} & 3.35 & \textbf{1.00} & 3.88 & 1.51 \\
Class ratio changes &  4.88 & 2.17 & 3.59 & \textbf{1.00} & 3.36 & 0.77 \\
Sub-cluster merge &  \textbf{2.67} & \textbf{3.67} & \textbf{2.67} & \textbf{1.00} & 5.00 & 2.68 \\
Sub-cluster move &  \textbf{3.17} & \textbf{3.17} & \textbf{2.67} & \textbf{1.00} & 5.00 & 2.68 \\
Sub-cluster split & \textbf{2.50} & \textbf{2.50} & 4.00 & \textbf{1.00} & 5.00 & 2.68 \\
Borderline examples &  \textbf{2.70} & 3.80 & \textbf{2.50} & \textbf{1.00} & 5.00 & 2.01 \\
Rare examples & \textbf{3.90} & \textbf{3.80} & \textbf{2.80} & \textbf{2.10} & \textbf{2.40} & 2.01 \\
\bottomrule
\end{tabular}
\caption{Average ranks in the Friedman test for different single factors in streams (the smaller, the better) -- where classifiers evaluated by Recall metric}\label{Tab:SingleDriftFriedman}
\end{table}

\begin{figure}[ht]
    \centering
    \subfloat{\includegraphics[width=5.7cm]{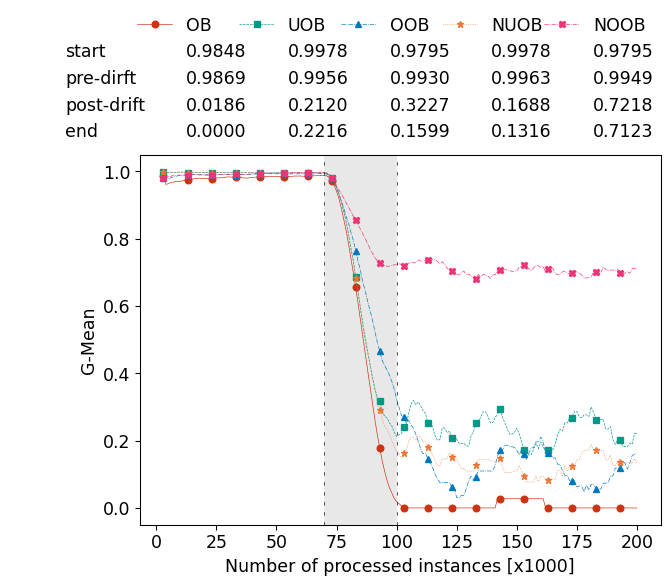}}
    \qquad
    \subfloat{\includegraphics[width=5.7cm]{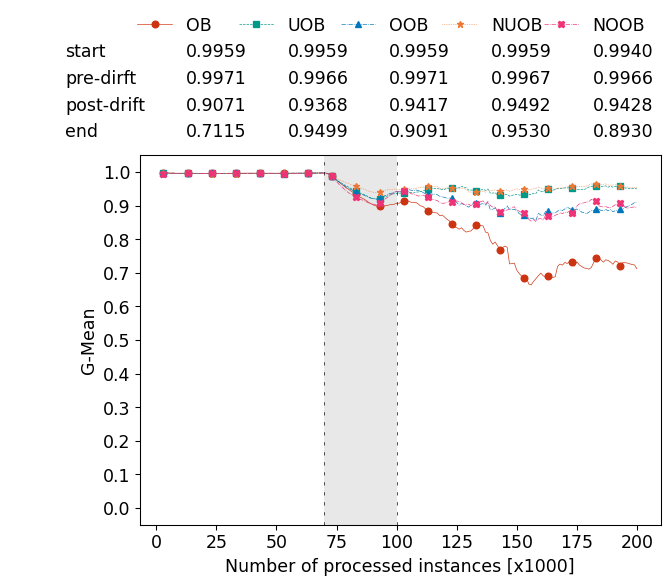}}
    \caption{Plots showing G-mean measure of bagging variants reacting to two kinds of drift \textit{80\% rare minority examples and Split 5 and imbalanced ratio changing from 10\% to 1\%} (the left-hand figure) and \textit{imbalance ratio 10\% and 80\% borderline minority examples}  (the right-hand figure)}
\label{Fig:pair}    
\end{figure}

Similar good performance of both proposed NUOB and NOOB classifiers was noticed for data streams with \textbf{pairs of factors}, in particular, if one of them was the presence of rare minority examples. An illustration of good reaction of NOOB to such drifts is presented in Figure \ref{Fig:pair} (the right-hand plot). 

\begin{figure}[ht]
    \centering
    \subfloat{\includegraphics[width=5.7cm]{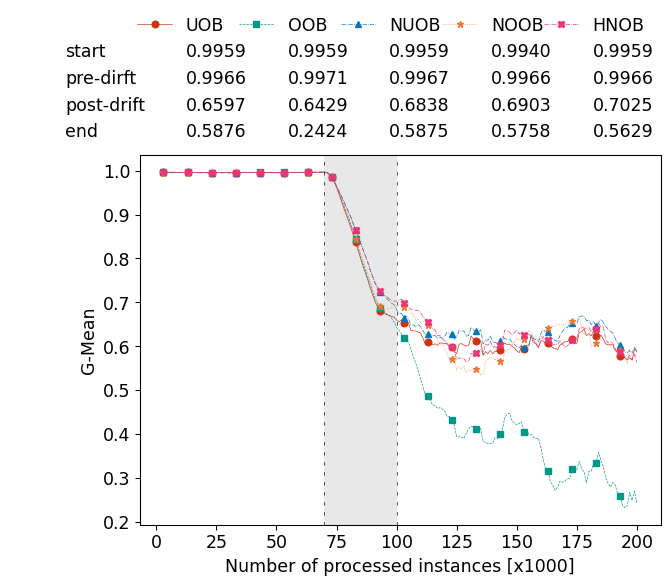}}
    \qquad
    \subfloat{\includegraphics[width=5.7cm]{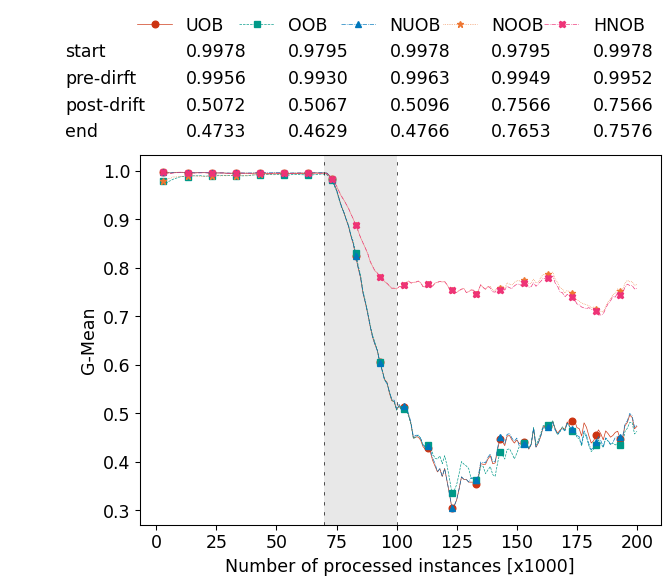}}
    \caption{Plots of  \textit{G-mean} measures for complex data streams \textit{Split5+Im1+Borderline40+Rare40} (the left-hand figure) and  \textit{StaticIm10+Im1+Rare80} (the right-hand figure)}
\label{Fig:multiple}    
\end{figure}

The usefulness of the new proposals is even more visible for the most difficult scenarios of generating \textbf{complex data streams} involving at least \textbf{three factors} occurring simultaneously. An example of such a reaction is shown in Figure \ref{Fig:pair}  (the left-hand sub-plot) for  \textit{StaticIm10+Im1+Rare80+Split5} combined factors. The global performance for different scenarios of such multiple factors is presented in Table \ref{Tab:ComplexFriedman}.

\begin{table}[ht]
\centering\small%
\setlength{\tabcolsep}{10pt} 
\renewcommand{\arraystretch}{1.5} 
\begin{tabular}{l l c c c c c c}
\toprule
Factors & Metric & OB & UOB & OOB & NUOB & NOOB & CD \\
\midrule
Multiple & G-mean & 4.60 & 2.69 & 3.27 & \textbf{2.30} & \textbf{2.14} & 0.43 \\
All  & & 4.39 & \textbf{2.56} & 2.99 & \textbf{2.58} & \textbf{2.48} & 0.31 \\
Multiple & Recall & 4.67 & 2.61 & 3.76 & \textbf{1.79} & \textbf{2.17} & 0.43\\
All  & & 4.58 & 2.60 & 3.62 & \textbf{1.53} & 2.67 & 0.31 \\
\bottomrule
\end{tabular}
\caption{Average ranks in Friedman test for multiple factors scenarios vs all examined factors}\label{Tab:ComplexFriedman}
\end{table}

Finally, the efficacy of Hybrid Neighbourhood Online Bagging was examined. The results showed its advantage over the best online bagging ensembles, in particular NUOB. Its good effectiveness was noticeable, especially for very complex scenarios involving \textbf{many factors} in the streams. An example of such a reaction is shown in Figure \ref{Fig:multiple} for \textit{StaticIm10+Im1+Rare80} and \textit{Split5+Im1+Borderline40+Rare40} combined factors.

\begin{table}[ht]
\centering\small%
\setlength{\tabcolsep}{10pt} 
\renewcommand{\arraystretch}{1.4} 
\begin{tabular}{l l c c c c c c}
\toprule
Factors & Metric & HNOB & UOB & OOB & NUOB & NOOB & CD \\
\midrule
Multiple & G-mean & \textbf{1.87} & 3.31 & 4.16 & 2.86 & 2.80 & 0.43 \\
All  & & \textbf{2.09} & 3.02 & 3.76 & 3.03 & 3.10 & 0.31 \\
Multiple & Recall & \textbf{2.25} & 3.15 & 4.63 & \textbf{2.32} & \textbf{2.66} & 0.43\\
All  & & 2.41 & 3.04 & 4.42 & \textbf{1.89} & 3.23 & 0.31 \\
\bottomrule
\end{tabular}
\caption{Comparison of Hybrid Neighbourhood  Online Bagging vs. other online bagging variants. Average ranks in the Friedman test}\label{Tab:ComplexFriedmanHNOB}
\end{table}

\begin{table}[ht]
\centering\small%
\setlength{\tabcolsep}{9.3pt} 
\renewcommand{\arraystretch}{1.5} 
\begin{tabular}{l c c c c c c}
\toprule
Pairs of factors &  HNOB & UOB & OOB & NUOB & NOOB & CD \\
\midrule
Imbalance+Move & \textbf{2.50} & \textbf{2.50} & 4.83 & \textbf{1.58} & 3.58 & 1.82 \\
Imbalance+Merge  & \textbf{2.42} & \textbf{2.32} & 4.83 & \textbf{1.67} & 3.75 & 1.82 \\
Imbalance+Split  & \textbf{2.39} & \textbf{2.56} & 4.83 & \textbf{1.56} & 3.67 & 1.47 \\
Imbalance+Borderl.  & \textbf{2.20} & 3.10 & 4.88 & \textbf{1.75} & 3.08 & 0.97 \\
Imbalance+Rare  & \textbf{1.95} & 3.58 & 4.83 & 2.90 & \textbf{1.75} & 0.97 \\
Split+Borderline  & \textbf{2.46} & 3.00 & 4.48 & \textbf{2.04} & 3.02 & 0.87 \\
Split+Rare  & \textbf{2.14} & 3.28 & 4.56 & \textbf{2.76} & \textbf{2.26} & 0.87 \\
\bottomrule
\end{tabular}
\caption{Average ranks in the Friedman test for different pairs of factors in streams
(the smaller, the better) – where classifiers evaluated by Recall metric}\label{Tab:DoubleDriftFriedmanHNOB}
\end{table}

Furthermore, we summarize in Table \ref{Tab:ComplexFriedmanHNOB} results of the Friedman test for multiple vs. all studied factors in synthetic data streams. As one can notice Hybrid Neighbourhood  Online Bagging is the best performing classifier. For the G-mean measure and multiple factors, his performance clearly outperformed all compared variants of bagging (the post hoc analysis and critical difference support for it). For the Recall measure, it is also the winner however due to the post-hoc analysis (CD = 0.43) its advantage over both NUOB and NOOB (also new proposed ensembles) is not so significant.

This trend was also observed in the Recall measure when analyzing streams with pairs of difficulty factors (see Table \ref{Tab:DoubleDriftFriedmanHNOB}). The proposed hybrid algorithm is the best in all of the analyzed scenarios but its superiority over other ensembles is not so significant, as confirmed by the post-hoc analysis.

\section{Discussion and Final Remarks}

As the empirical studies have shown that previously known streaming online classifiers cannot cope well with imbalanced data streams affected by difficulty factors such as the presence or drift of unsafe minority types of examples (including borderline or rare examples), in this paper we have introduced specialized extensions of resampling online bagging, namely Neighbourhood Undersampling or Oversampling Online Bagging ensembles. The main element of their design is to use the safety assessment of the currently processed example based on the distribution of labels of the other examples in its local neighbourhood. This leads us to an appropriate modification of the degree of transfer of these processed examples to component classifiers in accordance with the Poisson distribution parameter. Depending on tuning an additional exponent $\Psi$ parameter it is possible to more or less focus on the most difficult examples.

The results of experiments conducted on many synthetic data streams, modeling various difficulty factors or drifts, showed that both introduced online bagging generalizations are quite effective. In particular, in line with the hypotheses made during their construction, they were most effective for scenarios with drifts in the types of examples that previous online re-sampling bagging could not cope with. Neighbourhood  Undersampling  Online Bagging is the best classifier for dealing with the presence of borderline examples - which could be explained by their internal mechanism of reducing the probability of learning from the most difficult examples and performing "cleaning" of the borderline region between classes. On the other hand, Neighbourhood Oversampling Online Bagging was sometimes better for dealing with rare examples - recall here that $\Psi$ parameter was tuned to a higher value increasing the chance of copying many such examples. In general Neighbourhood  Undersampling  Online Bagging was the best classifier in most complex scenarios including multiple difficulty factors or their drift. However, the best choice is to use Hybrid Neighbourhood  Online Bagging which dynamically uses the currently superior model of a pair of parallel trained online bagging ensembles. 

Despite its definitely best experimental results (see e.g. Table \ref{Tab:ComplexFriedmanHNOB}), it opens the way for further research on more advanced ensembles of classification models and on stronger modifications of their structure as a result of detected drifts.

\subsubsection{Acknowledgements} The work of the second authors has been partly supported by PUT internal grant no. SBAD/0740 and (the second author) by the National Science Centre, Poland (Grant No. 2023/51/B/ST6/00545).

%
%
%
\bibliographystyle{splncs04}
\bibliography{mybibliography}
%


\end{document}